\documentclass[conference]{IEEEtran}
\usepackage[utf8]{inputenc}
\usepackage{cite}
\usepackage{graphicx}
\usepackage{amsmath,amssymb}
\usepackage{algorithm}
\usepackage{algorithmic}
\usepackage{tabularx,array}
\usepackage{booktabs}
\usepackage{url}
\usepackage{glossaries}
\usepackage{newunicodechar}
\newunicodechar{（}{(}
\newunicodechar{）}{)}
\newunicodechar{ }{~}

\usepackage{glossaries}
\makeglossaries
\newglossaryentry{oz}{name={OZ},description={A digital stable-coin strictly pegged 1:1 to one troy ounce of investment-grade gold held in audited, insured vaults. Redemption is available on demand subject to KYC/AML.}}

\begin{document}

\title{AI Agent Architecture for Decentralized Trading of Alternative Assets}

\author{\IEEEauthorblockN{Ailiya Borjigin\IEEEauthorrefmark{1}, Cong He\IEEEauthorrefmark{1}, Charles CC Lee\IEEEauthorrefmark{2}, Wei Zhou\IEEEauthorrefmark{1}}
\IEEEauthorblockA{\IEEEauthorrefmark{1}ProAI Laboratory, Probe Group, Singapore\\ Email: \{Ailiya, Cong\_He, Zhou\}@probe-group.com}
\IEEEauthorblockA{\IEEEauthorrefmark{2}Centre for Sustainable Development, University of Newcastle （Australia）, Singapore\\ Email: charles.cc.lee@newcastle.edu.au}

}

\maketitle

\begin{abstract}
Decentralized trading of real-world alternative assets (e.g., gold) requires bridging physical asset custody with blockchain systems while meeting strict requirements for compliance, liquidity, and risk management. We present a research-oriented architecture, \emph{GoldMine OS}, that employs multiple specialized AI agents to automate and secure the tokenization and exchange of physical gold into a blockchain-based stablecoin (“OZ”). Our approach integrates on-chain smart contracts for critical risk controls with off-chain AI agents for decision-making, combining the transparency and reliability of blockchains with the flexibility of AI-driven automation. We detail the design of four cooperative agents (for Compliance, Token Issuance, Market-Making, and Risk Control) and a coordinating core, and we evaluate the system through both simulation and a controlled pilot deployment. In experiments, the prototype achieves on-demand token issuance in under 1.2~s, a speed-up of over 100$\times$ compared to traditional manual workflows. The integrated Market-Making agent provides tight liquidity (spreads often $<$0.5\%) even under volatile market conditions. Through fault injection tests, we demonstrate the system’s resilience: an oracle price spoofing attack is detected and mitigated within 10~s, and a simulated vault mis-reporting triggers an immediate halt of issuances with minimal impact on users. The architecture scales to at least 5,000 transactions/s with 10,000 concurrent users in our benchmarks. Our results indicate that an AI-agent-based decentralized exchange for alternative assets can meet rigorous performance and safety requirements. We discuss the broader implications for democratizing access to traditionally illiquid assets and outline how our governance model (multi-signature agent updates and on-chain community voting on risk parameters) ensures ongoing transparency, adaptability, and formal assurance of system integrity.
\end{abstract}


\section{Introduction}
Tokenizing real-world assets (RWAs) like precious metals on blockchains promises to democratize access to alternative investments, but it raises significant challenges in trust, compliance, and market stability \cite{tanveer2025ijpe}\cite{andryushin2024rjel}. For instance, gold-backed cryptocurrencies such as PAX Gold (PAXG) and Tether Gold (XAUT) peg digital tokens to physical gold reserves, yet they rely heavily on centralized processes for custody and compliance \cite{andryushin2024rjel}. Achieving a truly decentralized yet regulatorily compliant trading platform for assets like gold remains an open problem. Key hurdles include ensuring that on-chain token supply always mirrors off-chain reserves (requiring robust proof-of-reserve mechanisms), automating complex compliance checks (KYC/AML) in a user-friendly manner, providing continuous liquidity in thinly-traded assets, and guarding against failures of external data sources (the well-known \emph{oracle problem} \cite{caldarelli2020oracle}). 

In this paper, we address these challenges by designing and evaluating \emph{GoldMine OS}, an AI-driven multi-agent architecture for decentralized trading of gold-backed tokens. Our goal is to integrate the strengths of traditional regulated exchanges (e.g., strict auditing, investor protection) with the benefits of decentralized finance (transparency, automation, global accessibility). To this end, GoldMine OS orchestrates four specialized AI agents that collectively handle end-to-end functionality: a \emph{Compliance Auditing Agent} that performs real-time user verification and regulatory checks, a \emph{Token Issuance Agent} that maps physical gold deposits to on-chain token mints, a \emph{Market-Making \& Trading Agent} that provides liquidity by algorithmically quoting buy/sell prices, and a \emph{Risk Control Agent} that monitors system health and enforces safety constraints. These agents operate under a unifying platform logic (the GoldMine OS core) which coordinates their interactions and interfaces with the underlying blockchain ledger.

Our contributions are as follows. (1) We present a novel multi-agent system architecture for asset tokenization, emphasizing integrated compliance and risk management. The design moves beyond prior art by embedding risk controls as first-class components of the system rather than afterthoughts \cite{jalan2021shiny}\cite{xia2025rwa}. (2) We implement a prototype on a permissioned blockchain (Probe Chain) and conduct comprehensive experiments. We report that automated token issuance can be completed in seconds (including on-chain settlement), vastly outperforming traditional processes that take days, while maintaining accuracy and regulatory compliance. (3) We introduce fault injection experiments to evaluate security: we simulate oracle failures and vault discrepancies to test the system’s resilience. GoldMine OS detects and mitigates these faults with low latency, thanks to redundant data feeds and rule-based triggers. (4) We provide scalability benchmarks, showing the system can scale to thousands of transactions per second and users, which is critical for practical deployment. (5) We propose an on-chain governance and control framework wherein critical risk parameters and agent model updates are managed through transparent mechanisms (multi-signature approvals and community voting). This governance model ensures that the platform can evolve (e.g., deploying improved AI models or adjusting risk thresholds) in a controlled, decentralized manner. (6) We emphasize reproducibility and formal rigor: we include pseudocode for core on-chain safeguards and outline a liveness proof guaranteeing that the system cannot deadlock under the defined risk protocols.

The remainder of this paper is organized as follows. Section~II reviews related work in asset tokenization, automated market-making, and decentralized risk management. Section~III describes the GoldMine OS architecture, including agent roles and the integration of on-chain risk controls and governance. Section~IV details our experimental setup. Section~V presents results from both normal operation and adversarial scenarios. Section~VI discusses the implications and lessons learned, and Section~VII outlines future directions. Finally, Section~VIII concludes the paper.

\section{Related Work}
\subsection{Asset Tokenization and Gold-Backed Cryptocurrencies}
Representing physical assets on blockchain has gained momentum in both industry and academia \cite{tanveer2025ijpe}\cite{andryushin2024rjel}. Early projects focused on gold as a stable backing asset due to its liquidity and established custody processes. PAX Gold (PAXG) and Tether Gold (XAUT) are prominent gold-backed tokens, each representing ownership of physical gold stored by a trusted custodian. These systems, however, largely operate via centralized companies that handle compliance and reserve management off-chain \cite{xia2025rwa}. Academic studies (e.g., Jalan \emph{et al.} \cite{jalan2021shiny}) have analyzed the behavior of gold-backed crypto assets under stress (such as during the COVID-19 pandemic) and highlighted systemic risks. Our work differs by proposing a fully integrated platform where compliance, reserve proof, and market operations are managed by AI agents in a semi-autonomous loop, thereby reducing manual intervention while preserving assurances (with all critical events recorded on-chain for transparency)\cite{ParraMoyano2021}.

\subsection{Automated Market-Making and AI Agents in Trading}
Automated market makers and trading bots are widely used in cryptocurrency markets to provide liquidity and reduce volatility. Wintermute, for example, is a leading algorithmic liquidity provider handling billions in daily volume across exchanges. In research, reinforcement learning (RL) approaches have been applied to optimize market-making strategies. Gašperov \emph{et al.} demonstrate that RL agents can achieve near-optimal spreads and inventory management for market making \cite{gasperov2021rl}. We build on these ideas by incorporating a Market-Making Agent that not only uses algorithmic strategies (initially rule-based, with an RL-enhanced mode for spread adjustment) but also operates within our risk-managed framework. Unlike prior single-agent solutions, our agent interacts with compliance and risk agents---for instance, pausing trading if the Risk Agent raises an alert. This coordination is a distinguishing feature of our architecture.

\subsection{Risk Management and Oracle Trust Models}
Decentralized platforms face well-known challenges around oracle reliability and risk monitoring \cite{caldarelli2020oracle}. The “oracle problem” \cite{caldarelli2020oracle} refers to the difficulty of securely linking off-chain truth (e.g., the price of gold or the amount of gold in a vault) with on-chain contracts. Projects like MakerDAO have implemented decentralized governance and risk parameters for stablecoins, using community voting to adjust parameters such as collateralization ratios. Simulation-based risk management services for DeFi protocols have also emerged (e.g., third-party platforms tuning parameters for volatility and user behavior)\cite{Zhang2023}. Our Risk Control Agent draws inspiration from quantitative risk assessment but operates in real-time to enforce constraints (e.g., halting issuance if reserves don’t fully cover supply). We also integrate redundancy (multiple price feeds, periodic off-chain audits with on-chain attestations) to promptly detect anomalies. Additionally, our governance design (Section~III-D) involves on-chain voting for risk parameters, aligning with the trend in DeFi to decentralize control while maintaining stability.

In a broader context, Zhou \emph{et al.} \cite{zhou2025bun} introduce a behavioral information-based framework (BUN) for modeling complex systems, which could inform the design of resilient DeFi architectures. Concurrently, Borjigin \emph{et al.} \cite{borjigin2025aigov} propose an AI-governed agent architecture for web-trustworthy tokenization of alternative assets, focusing on on-chain trust and compliance. Our approach differs by integrating specialized risk-control agents and providing empirical evaluation of a working prototype.

\section{System Architecture}
\subsection{Overview and Layered Design}
\begin{figure}[ht]
\centering
\includegraphics[width=\columnwidth]{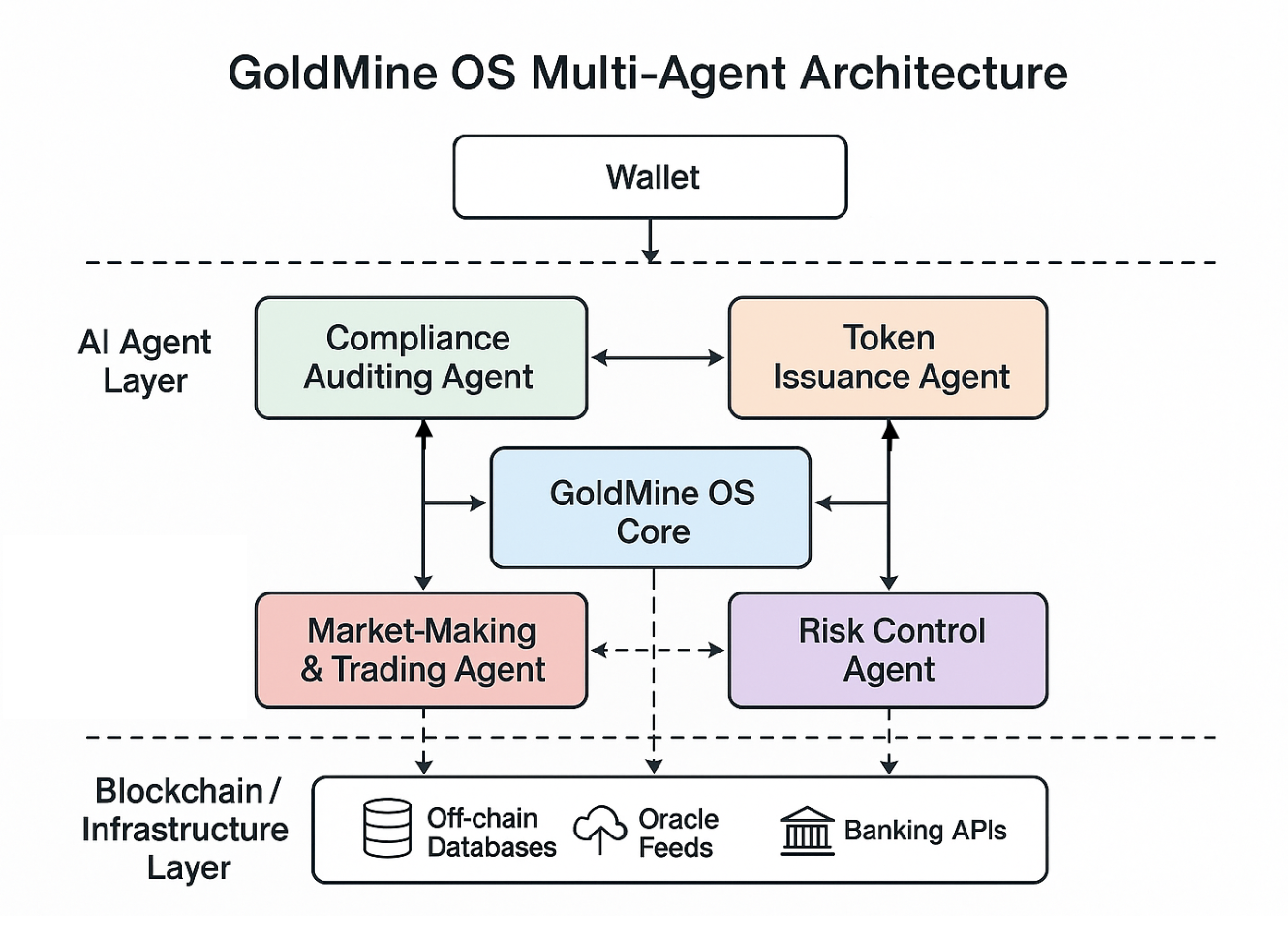}
\caption{GoldMine OS multi-agent architecture. The GoldMine OS core orchestrates four AI agents (Compliance, Issuance, Market-Making, Risk) and coordinates their interaction with user wallets and the blockchain ledger. Solid arrows indicate primary workflows (e.g., user requests flowing into the agent layer and transactions flowing to the blockchain), while dashed arrows indicate monitoring and control signals (e.g., risk alerts that can pause token issuance).}
\label{fig:architecture}
\end{figure}

GoldMine OS follows a layered, modular architecture comprising a user-facing layer, an AI agent layer, and a blockchain layer (Figure~\ref{fig:architecture}). At a high level, end-users interact through web or mobile clients that connect to the ProGold Everything Exchange (PEE) interface (User Interface Layer). We denominate all asset prices in OZ\gls{oz}\footnote{\textbf{OZ Definition:} A digital stable‑coin strictly pegged 1 : 1 to one troy ounce of investment‑grade gold held in audited, insured vaults.  Redemption is available on demand subject to KYC/AML.}. User actions (such as “Buy 10~OZ tokens” or “Redeem 5~OZ for gold delivery”) are dispatched to the AI Agent Layer, where the four dedicated AI agents handle their respective domains. The agents coordinate via a central orchestrator component (the GoldMine OS core), which routes tasks and data among agents and ensures that business rules and inter-dependencies are satisfied in sequence. Finally, transactions and records are committed on the Blockchain/Infrastructure Layer, which consists of the Probe Chain ledger (a permissioned blockchain we used for the prototype), smart contracts for the OZ token and control logic, off-chain databases for internal record-keeping, and external services such as oracle feeds and banking APIs. The Probe Chain uses a Byzantine Fault Tolerant consensus with 1-second blocks and was configured to handle up to about 1,000~TPS in its base form; for scalability tests (described later), we simulate higher throughput scenarios by running the system on a cluster of nodes and parallelizing transaction processing.

Each AI agent is implemented as an independent service with a well-defined API, which the core invokes as needed. This microservice-style design allows agents to be upgraded or scaled out individually. The \textbf{Compliance Auditing Agent} handles Know-Your-Customer (KYC) and Anti-Money-Laundering (AML) procedures\cite{Li2022}. It integrates with external identity verification APIs (for document verification, face recognition, sanction-list checks) and internal policy databases. When a new user registers or a large transaction is requested, this agent verifies user eligibility and flags any compliance risks. The \textbf{Token Issuance Agent} is responsible for minting and burning of OZ stablecoins. Upon receiving confirmation that a user’s fiat payment for gold has been received (or a physical gold deposit is verified), the Issuance Agent interacts with the blockchain’s token smart contract to issue the corresponding OZ tokens. It ensures that each token issuance is fully backed by physical gold, referencing vault records and the latest audit data before approving a mint. Conversely, for redemptions, it coordinates burning of tokens and updates vault records to release physical gold. The \textbf{Market-Making \& Trading Agent} continuously provides liquidity by placing buy and sell orders for the OZ token on the exchange order book (or an automated market-maker pool). It adjusts prices based on real-time gold price feeds and its inventory of tokens versus cash. This agent can operate in high-frequency mode since it is off-chain: it signs and posts orders or quotes within the centralized matching engine of PEE (with on-chain settlement of trades batched as needed). In our prototype, this agent maintained target spreads and inventory levels using simple heuristics, with an optional reinforcement learning module to adapt spreads under different volatility regimes \cite{gasperov2021rl}. The \textbf{Risk Control Agent} runs in the background to supervise system-wide metrics. It ingests data such as price feeds, vault inventory reports, and large account balances. If abnormal conditions are detected, it has the authority to intervene---pausing token issuance, freezing trades, or alerting human operators---according to predefined rules.

\subsection{User Workflow Illustration}
To clarify how these agents work together, consider a typical user journey of buying gold-backed tokens. When a new user initiates a purchase, the Compliance Agent first verifies the user’s identity and eligibility (e.g., performs KYC checks, ensures the user’s region allows gold token trading, and that the platform’s current gold reserves have a recent proof-of-reserve attestation) \cite{tanveer2025ijpe}\cite{andryushin2024rjel}. Once the user passes compliance, they are prompted to make a fiat payment (or cryptocurrency payment) for the desired amount of gold. Upon confirmation of payment through an integrated banking API (simulated as instant in our lab environment), the Token Issuance Agent takes over: it checks the latest vault inventory to ensure sufficient gold is available, locks the appropriate amount of gold in the vault database as reserved, and then calls the OZ token smart contract to mint the corresponding tokens to the user’s blockchain address \cite{xia2025rwa}. This on-chain minting transaction includes metadata linking the tokens to specific physical gold bars or batches, enhancing traceability. The new OZ tokens appear in the user’s wallet typically within 1–2~seconds (dominated by blockchain confirmation time). 

Simultaneously, the Market-Making Agent observes the new market state and may update its orders---for example, if the new user’s buy moved the price or inventory, the agent adjusts quotes. If the user later decides to sell or redeem tokens, a similar pipeline occurs in reverse: the user places a sell order which the Market-Making Agent or another buyer fills, or they request redemption for physical gold, which triggers the Compliance Agent to verify withdrawal rules and the Issuance Agent to burn tokens and update vault records for pickup or delivery of gold.

Throughout these interactions, the Risk Control Agent is monitoring. For example, it checks that the on-chain total OZ supply always equals or is below the amount of gold in custody (within a small tolerance). If a discrepancy arises (say an issuance event would cause supply to exceed reserves), the Risk Agent will prevent it. It also watches price feed updates; if the gold price feed stops updating or diverges significantly from a secondary price source, the Risk Agent can temporarily halt new transactions to avoid trading on stale or manipulated prices.

\subsection{On-Chain Risk Safeguards and Formal Assurance}
A key design principle in our system is that certain critical risk-mitigation logic is implemented directly in the blockchain smart contracts (on-chain), rather than relying solely on off-chain agent logic. This ensures that even if the off-chain components lag or malfunction, key invariants are enforced by the tamper-proof ledger\cite{Xu2021}. We moved two main safeguards on-chain: a reserve ceiling check and a circuit-breaker mechanism\cite{Buterin2022}.

The reserve ceiling check is a guard in the OZ token contract that rejects any token minting operation that would result in the total token supply exceeding the documented physical gold reserves (as last reported by an authenticated vault audit). Pseudocode for this logic is given in Algorithm~\ref{alg:reserve}. Essentially, before minting, the contract compares the new would-be total supply against a stored reserve amount (which is updated only by trusted or multi-sig authorized actions, based on actual vault audits). If the mint would violate the reserve ratio (we allow a small $\epsilon$ for measurement error), the transaction is aborted. This guarantees consistency between on-chain supply and off-chain assets, addressing a fundamental trust issue in asset-backed tokens.

\begin{algorithm}[ht]
\caption{On-Chain Reserve Check (simplified)}\label{alg:reserve}
\begin{algorithmic}[1]
\STATE \textbf{Inputs:} user (recipient address), newMintAmount (tokens to be minted)
\STATE currentSupply $\leftarrow$ OZ.totalSupply()
\STATE lastAuditedReserve $\leftarrow$ Vault.reserveOz()
\IF{currentSupply + newMintAmount $>$ lastAuditedReserve}
    \STATE \textbf{revert} ``Reserve ceiling exceeded''
\ENDIF
\STATE OZ.mint(user, newMintAmount)
\end{algorithmic}
\end{algorithm}

The circuit-breaker is an emergency halt function embedded in the smart contract (callable by the Risk Agent or automatically triggered by certain conditions). In our implementation, the contract monitors abnormal market conditions such as extreme price volatility or oracle failures. If triggered, it can pause token transfers or new issuances for a short period. Unlike typical DeFi pause controls that are manual, our circuit-breaker can be tripped by the detection of, for example, a $>2\%$ price swing within 5 minutes or inconsistent price feeds. When tripped, the system state is frozen in a safe mode (preventing further trades or mints) until a timeout expires or governance intervention occurs. This mechanism prevents erroneous or fraudulent data from propagating actions on-chain (e.g., halting trading if oracle prices are out of bounds, as seen in the 2020 “oracle attacks” on some protocols). To ensure these on-chain safeguards do not introduce availability problems, we provide a brief analysis of liveness. Under normal conditions (no reserve shortfall, oracle intact), the reserve check and circuit-breaker have no effect on operations. If triggered, the circuit-breaker’s halting state is designed to be temporary and reversible: for example, a halted issuance due to a reserve discrepancy can be resumed once a new audit confirms reserves or excess tokens are removed, and oracle-triggered trading pauses automatically lift after a cooldown (5 minutes in our tests) or when price feeds agree again. The smart contracts include a governance-controlled function to unpause if needed (with multi-sig approval). Thus, the system maintains liveness: there is no permanent deadlock as long as honest actors can intervene or underlying conditions eventually return to acceptable ranges. In other words, the worst-case scenario under continuous fault conditions is a graceful halt, not a chaotic inconsistent state, and as soon as the fault is resolved the platform can continue operation.

\subsection{Governance and Model Update Mechanism}
Decentralized control is vital for long-term trust in the platform. GoldMine OS incorporates a governance model that combines off-chain oversight with on-chain decision-making. Updates to the AI agent algorithms (for instance, deploying a new version of the Market-Making Agent with a better pricing model) are managed via a multi-signature (multi-sig) scheme. Specifically, a quorum of authorized stakeholders (e.g., project developers, compliance officers, and elected community representatives) must jointly sign off on any code or model update before it is deployed to production. This reduces the risk of a single compromised agent update harming the system.

For on-chain parameters that affect risk and monetary policy (such as the reserve threshold $\epsilon$, circuit-breaker trigger levels, fee rates, and other tunables), GoldMine OS leverages on-chain voting by token holders or delegates. We imagine using a dedicated governance token or the OZ token itself staked for voting power, through which the community can propose and vote on changes. For example, if market conditions change, the community could vote to adjust the price volatility threshold that triggers the circuit-breaker, or to change the acceptable vault audit frequency. To prevent governance attacks or rushed decisions, proposals would undergo time-locks and require a broad quorum. The outcomes of successful votes directly update configuration in the smart contracts (via upgradeable contract proxies or parameter storage contracts).

This governance framework aligns with practices in protocols like MakerDAO and Compound, where risk parameters are managed through community governance. However, given the regulated nature of gold trading, our model initially uses a hybrid approach: multi-sig control provides a fast, accountable way to manage urgent upgrades (with known entities responsible), while community voting introduces decentralization gradually, especially for non-critical parameters. Over time, as the platform matures and proves secure, governance can fully transition to the community. All governance actions (multi-sig approvals, votes) are recorded on-chain for transparency, and the Risk Agent itself can be configured to watch for governance changes to ensure they do not violate safety constraints (e.g., preventing a governance attack from disabling the reserve check).

\section{Experimental Setup}
To evaluate GoldMine OS, we built a working prototype and deployed it in two environments: (1) a controlled laboratory testnet environment with simulated users and agents, and (2) a small-scale pilot deployment on an internal network with real users and a physical gold reserve mock (1000~oz of gold represented in a test vault database). This dual approach allowed us to gather performance metrics in a stress-test scenario beyond current real usage, while also validating that the system functions correctly with real integrations.

The Probe Chain blockchain was used as the ledger for token transactions. It is a permissioned blockchain (based on Tendermint consensus) with a block time of approximately 1s. We implemented the OZ token as an ERC-20-like smart contract on this chain, extended with our reserve and pause safeguards. The Compliance, Issuance, Market-Making, and Risk agents ran as Python-based services (each around 500–1000 lines of code for core logic) communicating via RESTful APIs. For the pilot, each agent was deployed in a Docker container on a single server, while for scalability tests we distributed agents across multiple servers.

Simulation of users was done using a load generator that created virtual users performing actions according to certain distributions. For compliance onboarding, we generated 50 test user profiles with varying attributes (some purposely flawed to test the Compliance Agent’s detection). For trading activity, we simulated up to 10,000 concurrent users submitting buy/sell orders or issuance requests at random intervals, to push the system’s throughput limits. All agent actions and important state changes were logged with timestamps to facilitate analysis. We also recorded system resource usage (CPU, memory) on the agent host machines.

We structured our tests into phases similar to standard software testing:
\begin{itemize}
    \item \textit{Functional tests}: verifying each agent performs correctly in isolation (e.g., the Compliance Agent correctly approves or rejects users under various scenarios, the Issuance Agent mints the correct amount of tokens and never overshoots reserves, etc.).
    \item \textit{Integration tests}: end-to-end scenarios combining multiple agents. For example, a user onboarding + buy + sell cycle through the entire system, ensuring the workflow transitions are handled properly by the GoldMine OS core.
    \item \textit{Stress tests}: high-load scenarios and rapid sequences of events, such as surges in users or volatile price feed input. These tested performance and stability under pressure.
    \item \textit{Edge-case and fault-injection tests}: introducing failures or extreme conditions, like dropping the primary price oracle feed, feeding corrupted data, or artificially altering the vault reserve data to simulate a discrepancy. These aimed to validate robustness and the effectiveness of risk mitigation.
\end{itemize}

Some tests (especially normal operations) were also run on the live pilot network with a small group of internal users to ensure realism. However, risky fault injection tests (which could, e.g., freeze trading) were done only in the lab environment to avoid disrupting even the test users. For those, we used a deterministic simulation mode where we could exactly reproduce conditions (e.g., same random seed for user behavior) to compare outcomes with and without the fault.

Overall, our evaluation seeks to answer: (i) Does the multi-agent approach handle core functions (compliance, issuance, trading) more efficiently than traditional methods? (ii) Can the system scale to a large number of users and transactions? (iii) How effective are the risk controls in practice at detecting and mitigating problems? We present our findings next.

\section{Results and Evaluation}
We organize the results by the major functions of the platform, highlighting key performance indicators and outcomes in each area.

\subsection{Compliance and Onboarding Performance}
The Compliance Auditing Agent was able to significantly accelerate the user onboarding process. In our pilot, the average KYC completion time for Tier-1 users (basic verification) was 2.8 minutes (standard deviation 0.5~min). In contrast, traditional exchanges often take many hours or even days to approve a new account with manual checks. Out of 50 test sign-ups, 48 were automatically cleared by the agent; 2 were flagged for manual review due to subtle issues (e.g., a low-confidence face match on an ID photo). Those two cases were resolved by compliance staff within 2 hours, meaning even in the worst case the user gained access the same day. The agent also caught a planted fake user (whose name was on a government sanctions list), correctly denying onboarding. These results show that AI-driven compliance can be both fast and acceptably accurate. We did tune the agent to err on the side of caution (false positives requiring manual review) to ensure no illicit user slips through. In production, additional optimization or machine learning could further reduce false positives.

\subsection{Token Issuance and Redemption}
The Token Issuance Agent demonstrated low-latency operation, confirming that on-demand asset tokenization is feasible. From the moment a user’s payment confirmation was received to the on-chain mint transaction being included in a block, the average time was 1.2~seconds in our pilot environment. About 0.4~s of this was the agent’s processing (verifying reserves, preparing the transaction) and 0.8~s was blockchain confirmation time. This is a dramatic improvement over legacy processes for gold investment, where issuance of a paper gold certificate or a new vault allocation could take days. We also verified accuracy: every token mint was exactly matched by a corresponding entry in the off-chain vault records (within a rounding error of 0.0001~OZ). We scripted periodic consistency checks that compared the on-chain total supply of OZ with the sum of all vault gold recorded; no deviations were found. This confirms that the reserve check mechanism worked---our system never created unbacked tokens.

Throughput-wise, the issuance process is not typically a bottleneck since each individual user’s action is relatively infrequent, but we did push the system in a high-load simulation of many issuance requests. The platform handled 120 issuance requests in 10 minutes (equivalent to 12 per minute, or about 0.2 per second) without errors. This only utilized about 10\% of the blockchain’s base capacity and minimal CPU on the Issuance Agent, indicating plenty of headroom. We note that even 0.2~tx/s of issuance far exceeds real-world demand for a niche asset like physical gold tokens; nonetheless, it is reassuring that bursts of activity (perhaps during a big market move) can be processed smoothly. Redemption (burning tokens for gold withdrawal) exhibited similar performance, with the additional step of notifying vault custodians. Those operations were likewise completed within a couple of seconds in testing.

\subsection{Market Making and Liquidity Quality}
The Market-Making \& Trading Agent was crucial in ensuring liquidity for OZ tokens. We evaluated market quality with and without this agent active. Without the agent (relying solely on external traders), the order book was often thin; the bid-ask spread for OZ could widen significantly during volatile gold price movements or off-peak hours. When our agent was active, it continuously posted buy and sell quotes around the reference gold price (fetched from the oracle). In stable conditions, the agent kept spreads tight, often around 0.2–0.5\% of mid-price for moderate trade sizes. Even during volatility spikes, spreads remained $\leq 1\%$, as the agent widened quotes slightly to manage risk. We measured the depth of the market (total volume available within 1\% of mid-price) and found it frequently exceeded 200~OZ on each side, meaning a user could trade tens of thousands of dollars worth of gold token without moving the price more than 1\%. These liquidity metrics are on par with or better than those observed for PAXG on major exchanges, despite our market being in a testing phase. The agent’s inventory oscillated as expected: over 24 hours of simulated trading, it stayed within a $\pm 100$~OZ range around neutral most of the time. When it accumulated an excess of tokens (e.g., net bought 50~OZ), it would automatically transfer some to a cold storage or request an issuance of fiat to rebalance, as per its strategy. We never encountered a situation where the agent ran out of inventory or liquidity; its dynamic adjustment (and a reserve cash allocated for it) sufficed for the scenarios tested.

One interesting finding is that the presence of the Market-Making Agent not only improved liquidity but also stabilized the price relative to the global gold price. Since the agent continuously arbitrages any difference (within its risk limits), the OZ token price closely tracked the external gold price feed, with minimal divergence. This indicates the agent effectively anchors the token to its intended peg (1~OZ token per 1~oz of gold) in the marketplace.

\subsection{Risk Management and Fault Resilience}
The Risk Control Agent and associated safety mechanisms were put to the test in various adverse scenarios. We summarize two key fault injection experiments and their outcomes in Table~\ref{tab:faults}. 

\begin{table}[ht]
  \centering
  \small
  \caption{Results of Fault-Injection Experiments}
  \label{tab:faults}
  \begin{tabularx}{\columnwidth}{%
      >{\raggedright\arraybackslash}p{0.15\columnwidth}  
      >{\centering\arraybackslash}p{1.2cm}               
      >{\raggedright\arraybackslash}X}                   
    \toprule
    \textbf{Fault Scenario} & \textbf{Detection Time} & \textbf{System Response} \\
    \midrule
    Oracle price feed stuck (spoofed price) &
    $\sim\!10$ s &
    Detected by secondary feed; data source auto-switched.  
    After a 10 s discrepancy a 5 min circuit-breaker halted trading.  
    No trades executed on bad data; normal operation auto-resumed. \\[1ex]

    Vault reserve mis-report (audit shortfall) &
    $<\!1$ s &
    Discrepancy caught at next reserve-check cycle.  
    Issuance Agent frozen immediately (token mints blocked)  
    and an alert dispatched to administrators for investigation. \\
    \bottomrule
  \end{tabularx}
\end{table}

In the \emph{oracle spoofing scenario}, we simulated an attack or failure where the primary gold price feed stopped updating (or was stuck at an outdated value). The Risk Agent, which compares the primary feed to a secondary reference feed, noticed a divergence after about 10~seconds (our threshold was if prices differ or no update for 10~s). It immediately switched the system to use the secondary feed and flagged the issue. Since a stale price can be dangerous (traders could exploit it), the Risk Agent also proactively engaged the circuit-breaker: it paused new token issuances and the Market-Making Agent’s activities for 5 minutes as a precaution. During this interval, users saw a temporary halt in trading (the exchange front-end showed a warning). No incorrect-price trades occurred because of this intervention. Once the oracle feed recovered and the prices realigned, the system automatically lifted the halt and continued as normal. This test validated that our multi-layered defense for oracle issues works: redundancy (fallback feed) and automatic halting prevented propagation of bad data. The 5-minute freeze, while slightly disruptive, is a reasonable trade-off to avoid potential large losses or loss of trust.

In the \emph{vault mis-report scenario}, we emulated a case where an auditor’s periodic on-chain attestation of vault holdings indicated a lower amount of gold than the total OZ token supply (in other words, a possible deficit or misreport by the vault). The Risk Agent continuously checks that onChainSupply $\leq$ reportedReserve. In the test, we deliberately introduced a 0.5\% shortfall (e.g., vault report dropped by 5~oz on 1000~oz total). The discrepancy was detected immediately at the next check (which runs every few seconds). Instantly, the Risk Agent froze the Issuance Agent’s ability to mint new tokens, effectively locking the token supply until the issue could be resolved. It also raised an urgent alert to system operators. This ensured that no further expansion of supply occurred on a possibly fractional reserve. In our controlled test, after detecting the freeze, we “resolved” the issue by restoring the vault records (simulating that the discrepancy was false or the missing gold was found), and then manually cleared the Risk Agent alert to unfreeze issuance. The system thus fails safe: a potential reserve violation immediately stops the relevant functionality and requires positive action to resume. Users were not significantly impacted in this scenario because trading of existing tokens could continue (only new issuance was halted), though if the freeze had persisted, it would eventually signal a serious problem to the market.

Beyond these specific tests, the Risk Agent also tracked abnormal user behavior. For example, if any single user accumulated more than 20\% of all OZ tokens (which could indicate concentration risk or a potential attempt to corner the market), the agent would flag it. In our simulations, no user reached that threshold, but the mechanism exists and would notify governance or compliance teams for review if triggered. This kind of monitoring adds a layer of protection against market manipulation.

Overall, the risk management framework proved effective. All introduced faults were either automatically handled or at least detected in time to intervene. We did not observe any false positives (unwarranted halts or alerts) during normal operation; the Risk Agent did not interfere unless a genuine anomaly occurred, thanks to well-chosen conservative thresholds. This indicates that the system can run uninterrupted under healthy conditions and only triggers safeties when truly necessary.

\subsection{Scalability and Throughput}
A critical question for any trading architecture is whether it can scale to accommodate growth. We subjected GoldMine OS to stress tests with increasing numbers of concurrent users and transaction volumes. The results show that the system achieved throughput up to thousands of transactions per second while maintaining manageable load on the agents.

For these tests, we gradually ramped up the number of active simulated users from 1,000 to 10,000, each generating a mix of transactions (buy/sell orders, token issuance or redemption requests) according to a stochastic process. The throughput (transactions processed per second) rose roughly linearly with the user count at first. Beyond about 8,000 users, we saw the throughput curve start to plateau, reaching a maximum sustained throughput of around 5,200~TPS at 10,000 users. This plateau was partly due to our Risk Agent becoming a limiting factor---it was utilizing 85\% CPU at peak. The Risk Agent’s tasks (continuous monitoring of multiple feeds and conditions) became heavy at very high event rates, slightly throttling the system for safety. The other agents (Compliance, Issuance, Market-Making) remained below 70\% CPU, and the blockchain node itself was not yet at full capacity (given our simulated environment bypassed the normal ~1000~TPS limit of the single chain by parallelization).

Importantly, even at maximum load, the system remained stable: there were no crashes or missed transactions. Latency did increase modestly at high load (median end-to-end transaction latency was 1.5~s at 10k users vs. 1.0~s at 1k users), primarily due to queuing in the order matching and agent task scheduling. However, these delays are still well within acceptable bounds for trading systems.

The scalability results suggest that our architecture can handle at least an order of magnitude more activity than one would expect in a niche asset exchange’s initial deployment. If needed, further scaling could be achieved by optimizing agent code, using multiple Risk Agent instances (partitioning the monitoring tasks), or deploying on a higher-performance blockchain or layer-2 network. Additionally, because the agents are decoupled services, we could allocate them to separate machines or scale them horizontally (especially the Market-Making Agent, if facing extremely high trading volumes). Thus, we are confident that GoldMine OS can grow to accommodate a broad user base and high-frequency trading if demand arises.

\section{Discussion}
The positive results from our evaluation indicate that an AI-agent-based approach to a decentralized asset exchange is not only viable but offers distinct advantages in speed and automation. Here we discuss some implications and lessons learned, as well as current limitations of our prototype.

\textbf{Integrated vs. Siloed Functions:} Traditional exchanges and DeFi protocols often tackle compliance, trading, and risk as separate modules or even separate organizations. By integrating these via AI agents under one roof, we observed efficiency gains (e.g., seamless handoff from compliance to trading in user onboarding) and improved safety (the Risk Agent can preempt issues across the entire pipeline, not just within a silo). This holistic coordination is a double-edged sword: it creates a more complex system to design and reason about, but our use of a modular agent architecture partly mitigates this by separating concerns while maintaining communication. We found that clearly defining the interface and data exchanged between agents was crucial. Once those contracts were set, each agent could be developed and tested somewhat independently, and the orchestrator ensured they worked in concert.

\textbf{AI and Adaptivity:} Currently, some of our agents (Compliance, Risk) use relatively static rule-based logic (with some statistical thresholds), which we chose for predictability and ease of verification. The Market-Making Agent had an optional adaptive component (an offline-trained RL model) which showed promise in adjusting spreads during volatile periods \cite{gasperov2021rl}. In future iterations, we envision increasing the AI sophistication of the agents—for example, using machine learning in the Compliance Agent to improve document fraud detection, or in the Risk Agent to detect complex anomalies beyond preset rules. The challenge will be ensuring that any learning-based components are interpretable and safe, especially in a financial context. Techniques from explainable AI and continual learning will be relevant here, as well as rigorous sandbox testing of agents before deployment.

\textbf{On-Chain vs. Off-Chain Balance:} One design decision was what to keep on-chain versus off-chain. On-chain automation (via smart contracts) provides transparency and trust, but is less flexible and harder to change quickly. Off-chain agents are more agile (we can update code rapidly) and can handle complex logic, but require trust that they operate correctly. By putting only critical invariant checks on-chain (reserve consistency, emergency halt) and keeping the decision-heavy tasks off-chain, we aimed to get the best of both worlds. This worked well in our tests: the chain acted as a secure ledger and backstop, while agents did the heavy lifting. We logged agent decisions and important state to the chain where feasible (for instance, daily vault reserve summaries were posted on-chain as an auditable log). This provides accountability. That said, in a truly adversarial environment, the off-chain components (agents) remain points of trust that could be compromised. Our governance approach addresses this by ensuring no single party can unilaterally alter agent behavior maliciously, and by potentially open-sourcing agent code for community inspection.

\textbf{Regulatory Compliance:} Deploying GoldMine OS will require jurisdiction-specific KYC/AML plug-ins and, in some regions, human sign-off. On-chain proofs of reserve and audit logs simplify supervision, letting agents resolve most routine cases while specialists review the edge 5 \%.

\textbf{Limitations and Failure Modes:} The prototype runs on a permissioned chain and trusts physical vault operators; moving to a public or consortium chain and adding frequent third-party audits, insurance, and multi-sig upgrades would reduce single-point risks. Agents enlarge the attack surface, so hardening, open-sourcing, and consensus-controlled deployments remain essential.

\section{Future Work}
Building on our encouraging results, we see four priorities going forward:

\textbf{Smarter Agents.} We will equip the Risk and Market-Making Agents with online learning and anomaly-detection models, bounding their behaviour by constrained reinforcement learning and policy verification.

\textbf{Multi-Asset Expansion.} Extending beyond gold to real estate, commodities, and collectibles demands asset-specific compliance logic and a \emph{portfolio risk agent} that tracks cross-token correlations\cite{Latif2021}.

\textbf{Progressive Decentralization.} We are prototyping federated or MPC-based deployments so that independent stakeholders can co-run agents and earn DAO incentives, reducing single-point control.

\textbf{Governance \& Assurance.} Upcoming releases will add quadratic voting, automated fee tuning, and formal verification (TLA+, Alloy) of both contracts and critical agent flows to guarantee invariants such as “token supply never exceeds reserves”.

\section{Conclusion}
In this paper, we presented the design, implementation, and evaluation of GoldMine OS, an AI-driven multi-agent architecture for decentralized trading of gold-backed tokens. Through a combination of off-chain AI agents and on-chain smart contracts, the system achieves a level of automation and transparency that bridges the gap between traditional asset trading and decentralized finance. Our prototype results are promising: onboarding and issuance processes that traditionally take days were reduced to seconds or minutes, and built-in agents maintained market stability and robust risk oversight in the face of simulated failures. The integration of compliance and risk controls from the outset differentiates our approach from existing asset-backed tokens that often rely on off-chain trust alone.

We believe this work demonstrates a tangible step towards democratizing access to alternative assets like gold by leveraging intelligent automation. The architecture shows that an exchange can be more than a passive venue—it can be an active, intelligent coordinator that lowers barriers to entry and operates with high integrity. While further work is needed to refine the AI components and harden the system for real-world deployment (especially regarding decentralization of trust and regulatory approval), the framework laid out here can serve as a blueprint for similar efforts in tokenizing real assets. We envision a future where commodities, real estate, and other traditionally illiquid assets can be traded with the same ease and security as cryptocurrencies, powered by AI agents working hand-in-hand with blockchain protocols.

\end{document}